\documentclass[11pt]{article}

\usepackage[final]{acl}

\usepackage{times}
\usepackage{latexsym}

\usepackage[T1]{fontenc}

\usepackage[utf8]{inputenc}

\usepackage{microtype}

\usepackage{inconsolata}

\usepackage{graphicx}
\usepackage[most]{tcolorbox}
\usepackage{pifont}
\usepackage{enumitem}
\usepackage{listings}
\usepackage{capt-of}
\usepackage{dashrule}
\usepackage{makecell}
\usepackage{adjustbox}
\usepackage{tabularx}

\tcbset{
  promptbox/.style={
    colback=gray!5,
    colframe=black!60,
    colbacktitle=gray!20,
    coltitle=black,
    fonttitle=\bfseries,
    title={#1},
    fontupper=\ttfamily\small,
    boxrule=0.8pt,
    arc=2mm,
    boxsep=2pt,
    left=6pt,
    right=6pt,
    top=4pt,
    bottom=4pt,
    enhanced
  }
}

\usepackage{booktabs}
\usepackage{rotating}
\usepackage{multirow}
\usepackage[dvipsnames]{xcolor}
\usepackage{comment}

\newcommand{\impr}[1]{\rlap{\textbf{\textsuperscript{\textcolor{teal}{\fontsize{3.8}{4}\selectfont #1}}}}}
\newcommand{\decr}[1]{\rlap{\textsuperscript{\textcolor{red}{\fontsize{6}{8}\selectfont #1}}}}

\usepackage{hyperref}
\usepackage[capitalise,noabbrev,nameinlink]{cleveref}

\newcommand{\jun}[1]{\textcolor{red}{\bf\small [#1 --Jun]}}

\newcommand{\suhang}[1]{\textcolor{blue}{[SW: #1]}}

\usepackage{tcolorbox}
\usepackage{xcolor}
\usepackage{marginnote} 
\makeatletter

\title{
  \leavevmode
  \llap{\raisebox{-0.1\height}{\includegraphics[height=3em]{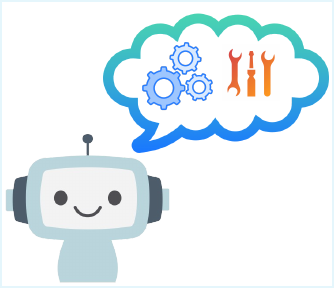}}\hspace{-1em}}%
  \textit{ToolDreamer}: Instilling LLM Reasoning Into Tool Retrievers
}

\author{
  \textbf{Saptarshi Sengupta\textsuperscript{1}\thanks{Work done during an internship at Bosch Research
North America.}},
  \textbf{Zhengyu Zhou\textsuperscript{2}},
  \textbf{Jun Araki\textsuperscript{2}},
  \textbf{Xingbo Wang\textsuperscript{2}},\\
  \textbf{Bingqing Wang\textsuperscript{2}},
  \textbf{Suhang Wang\textsuperscript{1}},
  \textbf{Zhe Feng\textsuperscript{2}},\\
  \texttt{\{sks6765,szw494\}@psu.edu}\\
  \texttt{\{zhengyu.zhou2,jun.araki,xingbo.wang,bingqing.wang,zhe.feng2\}@us.bosch.com}\\
  \textsuperscript{1}The Pennsylvania State University, 
  \textsuperscript{2}Bosch Research North America,
}

\begin{document}
\maketitle
\begin{abstract}
Tool calling has become increasingly popular for Large Language Models (LLMs). 
However, for large tool sets, the resulting tokens would exceed the LLM's context window limit, making it impossible to include every tool. Hence, an external retriever is used to provide LLMs with the most relevant tools for a query. Existing retrieval models rank tools based on the similarity between a user query and a tool description (TD). This leads to suboptimal retrieval as user requests are often poorly aligned with the language of TD.
To remedy the issue,
we propose \textit{ToolDreamer}, a framework that conditions retriever models to fetch tools based on \textit{hypothetical (synthetic)} TD generated using an LLM, i.e., descriptions of tools that the LLM feels will be potentially useful for the query. The framework enables a more natural alignment between queries and tools within the language space of TD's. We apply \textit{ToolDreamer} on the \texttt{ToolRet} dataset and show that our method improves the performance of sparse and dense retrievers with and without training, showcasing its flexibility. With our proposed framework, we aim to offload a portion of the reasoning burden to the retriever so that the LLM may effectively handle a large collection of tools without inundating its context window.

\end{abstract}

\section{Introduction}\label{sec:intro}

\begin{figure}
    \centering
    \includegraphics[width=\columnwidth]{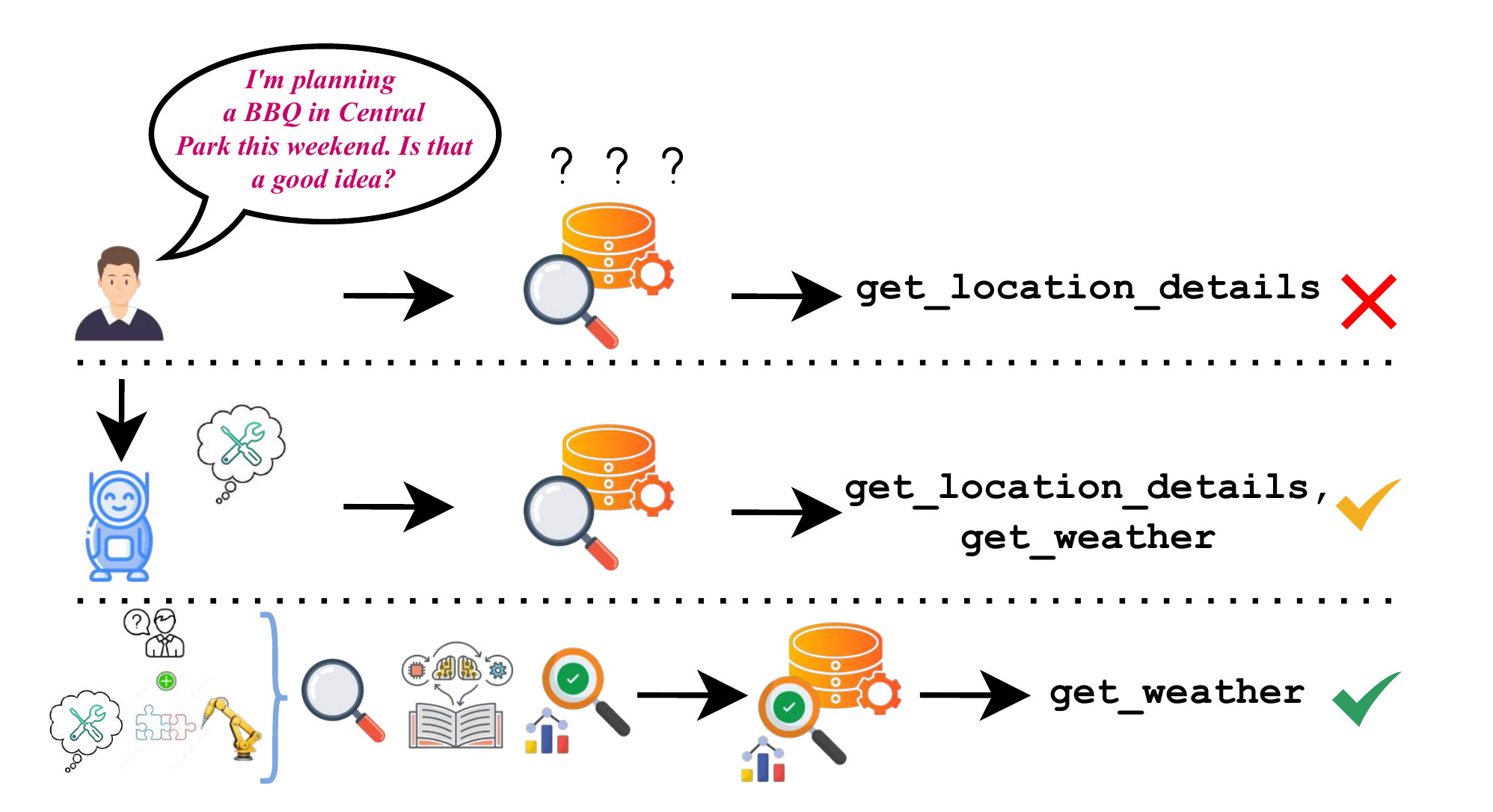}
    \caption{Illustration of the problem statement/motivation. Retrievers conditioned to learn query-tool relationships struggle to locate tools that are not readily obvious from the question (top). LLM-generated hypothetical tools can potentially be used to locate the right tool, but it still misses the mark as it has not been \textit{conditioned} to learn such tool-to-tool relationships (middle). Thus, if we first \textit{align} the hypothetical tool with its gold tool counterpart, and incorporate the query background to \textit{train} the retriever, it would lead to a far superior model capable of accurately locating the correct tool (bottom).}
    \label{fig:motivation}
\end{figure}

LLM-based \textit{tool calling} (generating text as parsable function calls) \cite{masterman2024landscape, qu2025tool} has emerged as a viable solution to real-world environment interaction, allowing them to solve a variety of complex tasks from mathematics \cite{das-etal-2024-mathsensei} to molecular structure generation \cite{zhang-etal-2025-automated} by using tools. 
To enable tool calling, an LLM is provided with the JSON schema of each tool, consisting of function arguments, data types, and their purpose \cite{huggingfaceToolUse}. Although tools are well-defined data structures, ultimately, for the LLM, they appear as additional tokens in its context window. When the number of tools is small, the entire set of tokens falls well within the model's processing limits. However, as the tool set expands, it puts pressure on the LLM, which ultimately leads to context overflow (out-of-memory (OOM)) (cf. App. \ref{sec:app_OOM}).

To address the OOM issue, LLMs are paired with external retrievers \cite{DBLP:journals/corr/abs-2502-14822} which aim to select the right tool(s) for a query. Traditionally, tool retrievers are trained on triplets (question, gold tool, negative tool) by optimising the InfoNCE contrastive loss \cite{oord2019representationlearningcontrastivepredictive, henderson2017efficient} between the three components. This means that retrieval models learn to maximise the similarity between the questions and their corresponding gold (required) tools and minimise the similarity between the questions and negative (irrelevant) tools \cite{shi-etal-2025-retrieval, 10.1145/3627673.3679847, xu-etal-2024-enhancing-tool}. However, we argue that this is suboptimal for tool retrieval and illustrate this in Fig. \ref{fig:motivation} (top). In the given query, there is no obvious indication for weather information, which is the most relevant here, considering a BBQ is an outdoor event strongly influenced by weather conditions. Retriever models rely on semantic similarity and cues from the provided text. As such, they are unable to perform the necessary reasoning needed to select the correct tool, and consequently return an incorrect tool when using the query directly. 

To alleviate the above issue, recently, \textit{hypothetical tool-generation} has been proposed, i.e., leveraging LLM-reasoning to generate tools, potentially useful for a query. In this regard, \citet{kachuee-etal-2025-improving} propose the first study wherein they utilise an LLM's commonsense knowledge \cite{patil2025advancing} to generate useful tool descriptions for retrieval. However, the issue here is that the downstream retriever is \textit{not optimised} for retrieval, as it does not undergo tool alignment training (learning to relate hypothetical tools to their gold counterpart). This defeats the purpose of generating such hypothetical tools, as the retriever is not conditioned to handle such hypothetical tool descriptions (Fig. \ref{fig:motivation} - middle).

Therefore, in this paper, we study a novel problem of retrieval optimization for retrieving tools based on hypothetical tools generated for a query. We see two challenges here: (i) data scarcity: considering the uniqueness of this problem, we do not find any existing datasets that enable training retrievers using question-hypothetical-gold tool samples, and (ii) knowledge infusion: determining the best way to integrate external knowledge as hypothetical tool descriptions into a retriever is a non-trivial task. To address these challenges, we propose a novel framework called \textit{ToolDreamer}. With \textit{ToolDreamer}, we first generate hypothetical tools for a question through LLM-prompting. These hypothetical tools are then \textit{aligned}/mapped to their gold counterpart, using bipartite graph matching, to form the training target. The aligned tools and their associated questions form the dataset used to train the retriever model, addressing the first issue. There are various ways to represent the input hypothetical tools to the retriever. \textit{ToolDreamer} couples questions with their aligned tools, by integrating LLM-generated hypothetical tool metadata, to train the retriever using an augmented InfoNCE loss that utilises both question and potential tools (\S \ref{sec:retriever_training}), thus tackling the second issue (Fig. \ref{fig:motivation} - bottom).

Our \textbf{main contributions} are: (i) We study an important and novel problem: optimizing tool retrieval by investigating query-tool v/s hypothetical tool - gold tool mapping; (ii) We present \textit{ToolDreamer}, a novel framework that can effectively retrieve tools based on hypothetical tools generated for a question; and (iii) Experimental results demonstrate the effectiveness of \textit{ToolDreamer}.

\section{Related Work}
\label{sec:related}


\textbf{Dynamic Tool Creation.} Recent years have seen the adoption of dynamic tool creation, i.e., using LLMs to synthesise new tools \cite{qian-etal-2023-creator, yuan2024craft, cai2024large, wolflein-etal-2025-llm}. This not only removes the need to write code for an extensive number of tools, but also allows for on-demand tool creation, i.e., it is not necessary to determine the entire tool-set a priori. However, the major issue with these studies is the complexity and cost involved in crafting robust code, which requires several rounds of iteration and even multi-agent processes \cite{wolflein-etal-2025-llm}. On the other hand, \textbf{our framework does not task an LLM to generate the actual implementation of the tool}, but rather descriptions of tools that can potentially be used to solve that query, which is a much simpler proposition for them.

\textbf{Tool Retriever Training.} Current approaches for training retrievers \cite{shi-etal-2025-retrieval, 10.1145/3627673.3679847, xu-etal-2024-enhancing-tool} optimise a contrastive loss objective between (query, gold tool, negative tool) triplets, which we argue is sub-optimal due to the aforementioned reasons (misalignment between tool descriptions and user queries). Our framework improves on these methods by using a better aligned anchor term, i.e., instead of training on query-tool pairs, we train on tool-tool pairs.


\textbf{Tool Retrieval With Auxiliary Information.} Recently, there has been an uptick in methods for generating better search requests for tool retrieval. \citet{chen-etal-2024-invoke} propose a \textit{training-free} framework where, instead of generating tools, they \textit{generate queries} from tool descriptions. These queries are then appended to the tool description to form an augmented target for test-time questions. While the tool set now contains additional useful information, it comes in the form of queries, which, as we have shown (Fig. \ref{fig:motivation}), is a suboptimal target for retrieval due to their nuanced nature. \citet{braunschweiler-etal-2025-toolreagt} propose a \textit{training-free} \textit{ReAct} \cite{yao2023react} agent that iteratively refines the query and tool set provided by the retriever. Similar to us, this framework uses an LLM to suggest potential tools for a query. However, they do not optimise their retriever using these potential tools, relying instead on several rounds of LLM reasoning, which can in turn accrue cost, particularly for API-based models. \citet{kachuee-etal-2025-improving} is the most similar to us in ideology. They leverage an LLM's commonsense knowledge to generate hypothetical tool descriptions for retrieval, and also train their LLM through SFT (Supervised Fine-Tuning) to generate high-quality tools. The key difference between our methods is that they train their LLM for tool-generation, using query-tool similarity (suboptimal as explained), and ignore retriever optimisation. Our proposed framework is inherently different from these works. As the retriever is not optimized for matching hypothetical tool descriptions to real tools, resulting in suboptimal retrieval performance, we propose a novel framework that fine-tunes the retriever and improves the performance with better retrieval query design.

\textbf{Instructional Retrieval.} A line of work related to the above is enhancing queries with additional instructions to guide retrieval. However, these works either use manually crafted \cite{weller-etal-2025-followir, sun-etal-2024-mair} or LLM-generated instructions \cite{shi-etal-2025-retrieval} to get retrieval directives. While more informative than the original query, these instructions may not match well with the tools space or fully decompose the query with tool description language, ending up with a suboptimal alignment in tool retrieval. In contrast, our approach (i) enforces the thought process of the LLM to better match the underlying meaning of the query with tools space (Fig. \ref{fig:motivation}: the weather tool is required while the query does not mention weather at all.), and (ii) appropriately decomposes a query to cover cases requiring multiple tools to answer (Fig. \ref{fig:generated_tool_ex}).

\section{The Proposed \textit{ToolDreamer}}

\begin{figure*}
    \centering
    \includegraphics[scale=0.34]{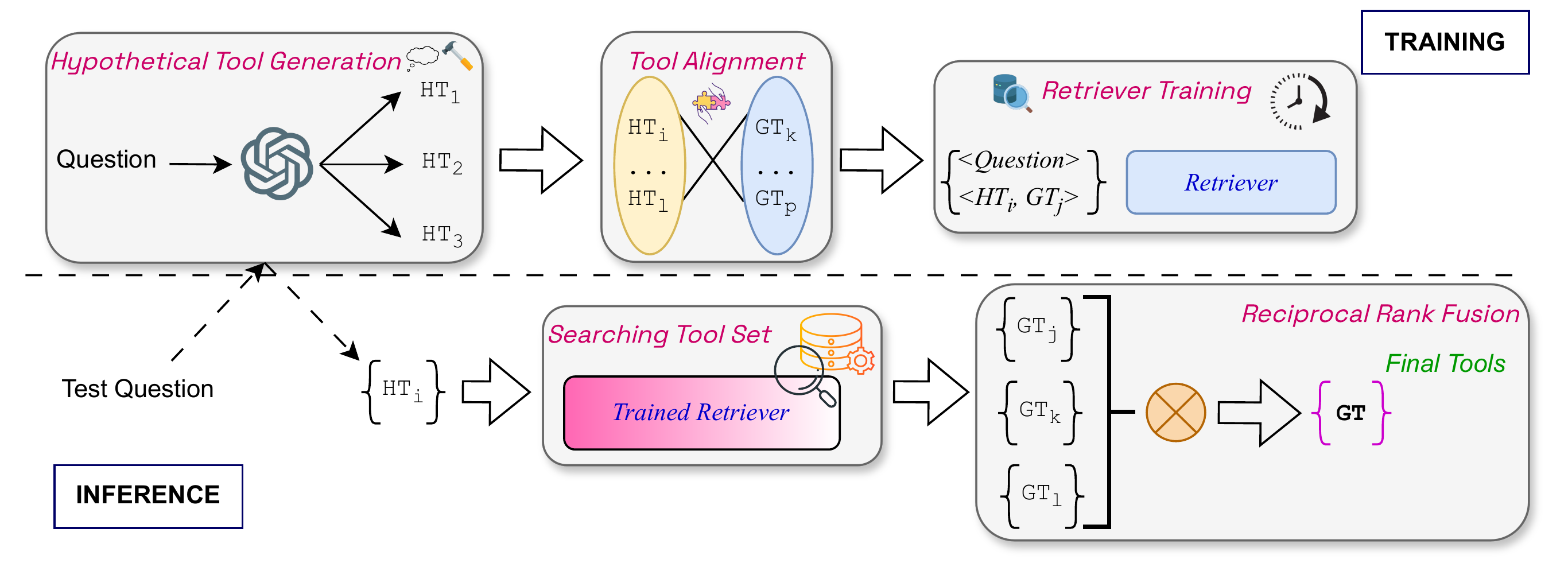}
    \caption{\textit{ToolDreamer} overview. \textbf{Training Phase (Top)}: A strong LLM first generates hypothetical tools (HTs) for a question. These tools are then aligned with their gold tool (GT) counterparts. Finally, the aligned tool pairs are used to train the retrieval model. \textbf{Inference Phase (Bottom)}: At test time, an unseen question is provided to the same LLM, which generates a set of hypothetical tools for it. For each tool, the trained retriever collects a top-K list of tools from the database. A unified list is obtained through reciprocal rank fusion (RRF).}
    \label{fig:framework}
\end{figure*}

In this section, we detail the inner workings of our framework, starting with tool retriever training, followed by the inference phase. An overview of the framework is given in Figure \ref{fig:framework}.

\subsection{Training Phase}
\label{sec:training_phase}


Our objective is to obtain a better retriever that is trained to understand user queries and relationships between LLM-generated hypothetical tools and gold tools for the associated query. However, the main challenge is obtaining the training data for the retriever, as we do not have the ground truth hypothetical-to-gold tool matching for each query. To address the challenge, we propose a novel two-step approach for \textit{training data generation}: given a question/query, an LLM generates potentially useful HT's. Next, we \textit{align} the generated HT's with their gold counterpart using a bipartite graph matching algorithm to form the triplets (\S \ref{sec:intro}). Though the alignment between HT and GT might not be perfect, it still gives a rough and sometimes accurate preference/matching between HT and GT, making it possible to train the retriever. We \textit{train our retriever} to assign high similarity between the aligned tools and low similarity between HT and randomly selected tools. This equips it with the ability to learn HT-GT relationships for a question and to perform a more targeted tool search.

\subsubsection{Hypothetical Tool Generation}
\label{sec:train_hyp_tool_gen}

This is the first step of the training phase. Here, given a question, we prompt an LLM to generate tools it thinks could be relevant for solving the question (See Fig. \ref{fig:ht_train_prompt} for Prompt Template). As we have access to the training data, we know a priori the number of tools required for the question. Thus, the LLM is instructed to generate hypothetical tools equal to the number of gold tools for the query. For the actual generation, we task the model to provide its thought process for the hypothetical tool, an appropriate name, and a description. The first two components are used because it has been shown \cite{yugeswardeenoo-etal-2024-question, zhu-etal-2025-deepreview} that LLM performance improves when providing a reasoning breakdown before the final answer. Additionally, as mentioned, we do not generate an implementation of the tool, but only metadata (thoughts + name + description). 

Although LLMs have gotten better over the years, they are still prone to making mistakes and not following instructions \cite{tong-etal-2024-llms, chen2024deep}. We notice a similar phenomenon in our study. While our LLM (\S \ref{sec:experiments}) can generate the right number of tools for the majority of questions, it falters for a very small fraction (0.3\%). These questions are subsequently removed from the sampled dataset. 

\subsubsection{Tool Alignment}


After generating each question's HT's, we align them with their gold counterpart. This is done to form the final training samples, where each question is augmented with its HT along with the target value (corresponding GT). First, we treat the hypothetical and gold tool sets as a bi-partite graph, i.e., a graph whose vertices can be partitioned into two distinct sets and edges always connect vertices from corresponding sets. We justify this assumption by noticing that the tools in each set belong to a distinct category (Hypothetical Tool (HT) v/s Gold Tool (GT)) and hence show no overlap apart from being related by a common theme, i.e., they are all tools. Thus, ``connecting'' tools can be viewed as a graph matching problem, i.e., finding a set of edges between the nodes such that no two edges have a common vertex. Simply put, obtaining edges between a unique pair of nodes.

We first compute a semantic similarity (using Qwen3-8B \cite{qwen3embedding} embedding) matrix for each (hypothetical tool, gold tool) pair. From this matrix, we perform minimal cost matching using the \textbf{Hungarian algorithm}\footnote{The Hungarian algorithm has a time complexity of $O(n^3)$ where $n$ is the number of elements (rows/columns of the cost matrix) to align. This can potentially introduce significant computation overhead into the framework. However, this holds only if the cost matrix to optimize is significantly large. In \texttt{ToolRet}, the number of tools required for a question ranges between 1 (56\% of questions) and 8 (0.1\%), with the average being 2. As such, their resultant matrices do not take considerable time to process.} \cite{https://doi.org/10.1002/nav.3800020109}. The algorithm solves the problem: given a task-cost matrix, what is the optimal way of assigning tasks to minimise overall cost? The assignment is done to create a \textbf{best possible} 1:1 mapping between elements. We leverage the same method to optimally connect tools based on their similarity and create our set of aligned (HT, GT) pairs. 

A natural concern here is \textit{misalignment} (\S \ref{sec:weak_tool_alignment}) (poor HT-GT match). As we deal with a square semantic similarity matrix, there will \textit{always} be a match. However, it is a \textit{best possible attempt} by the underlying matching algorithm and embedding model used to score HT-GT similarities. Noise can be greatly mitigated by using a strong embedding model and a reliable alignment method. As mentioned previously (\S \ref{sec:training_phase}), HT-GT matching might not be perfect. However, this imperfect mapping acts as a rough and often accurate proxy for the true alignment, making it possible to train the retriever. The ultimate goal is to train the retriever to assign high similarity between the aligned tools and low similarity between HT and random tools. We do this to make it perform a more targeted tool search, using the learned HT-GT relationships.

\subsubsection{Retriever Training}
\label{sec:retriever_training}

A retriever that has not been conditioned to learn HT-GT relationships will offer reduced benefits when using HT as a search vector (Fig. \ref{fig:motivation}). As such, it needs to be trained to identify the most relevant gold tool for a given question and its HT's. To train our retriever, we use the assembled training dataset, i.e., questions + aligned HT-GT. We optimise our retriever using InfoNCE loss. The basic idea is to teach the retriever to push the HT towards its GT, in embedding space, while simultaneously pushing the HT away from negative (irrelevant) tools.


Each HT is associated with three components, i.e., the thought process of the LLM behind suggesting that tool, the tool name, and its description. There are different ways to integrate this information (\S \ref{sec:intro}) into the retriever for selecting gold tools. We could use the HT name and description only as input to the retriever. However, the LLM-thoughts also contain important \textit{reasoning information} for suggesting the necessity of the generated HT, which provides an additional signal to the retriever. Thus, to take advantage of the entire metadata, we format our input to the retriever as, 

\vspace{8pt}
\fbox{\small \ttfamily \textbf{T}houghts:\{\} Tool \textbf{N}ame:\{\} Tool \textbf{D}escription:\{\}}
\vspace{8pt}

We denote this input style as \texttt{TND} (Thought-Name-Description). As the query also provides an important signal on what tools to retrieve, we formulate another input style where we prepend the query (\textit{Q}) to \texttt{TND}, denoting it as \texttt{QTND}. 

Overall, our final loss function is given in Eq. \ref{eq:infonce_loss}. $A$ = \texttt{TND} / \texttt{QTND} is the \textit{anchor}, i.e., the reference for which the model has to be optimised; $GT$ = Gold Tool; $n_{i}$ is the $i^{th}$ negative/irrelevant tool for a training sample, and \textit{sim} = semantic similarity function, usually cosine similarity. $[\cdot]$ represents the embedding for the corresponding term. We provide training/hardware details in App. \ref{sec:training_details}.  
\begin{equation}
     -\log\frac{e^{sim([A], [GT])}}{e^{sim([A], [GT])} + \sum_{i=1}^{k}e^{sim([A], [n_{i}])}}
     \label{eq:infonce_loss}
\end{equation}


\subsection{Inference Phase}

Our inference phase also proceeds in three steps. At first, we generate HT for the test data, similar to the training phase. Next, we use our trained retriever model to perform tool-retrieval using the improved search objective (question + HT). Finally, each retrieved top-K tool list, corresponding to each HT, is merged to form a unified result. 

\subsubsection{Hypothetical Tool Generation}

This step is similar to the training step, where an LLM (\S \ref{sec:experiments}) is asked (Fig. \ref{fig:ht_test_prompt}) to generate hypothetical tools for the given query. During inference, we do not know a priori how many or what tools to retrieve. Thus, we do not tell the LLM how many hypothetical tools to generate. As such, it is tasked with generating as many tools as it feels are relevant for the query in an open-ended manner.

\subsubsection{Tool Retrieval}

As explained in the training phase, we investigate two retrieval settings, viz., using \texttt{TND} (HT Thought-Name-Description) and \texttt{QTND} (Query + HT-\texttt{TND}). For each strategy, the respective trained retriever returns a top-K list of tools, using the search vectors (HT's) for a question. 


An LLM is not always able to follow instructions correctly. Consequently, for a very small fraction of questions, it does not generate an equal number of thoughts, names, and descriptions. As we require each HT to be associated with this metadata to form the final search vectors, to handle these instances, we ignore the HT generations and fall back on the base question as the search vector. Despite repeated prompt changes, the LLM was unable to generate the correct number of components for \textit{all} questions, i.e., it was found to always falter for \textit{some} questions. However, this is not a major issue as the number of such samples is not large enough to introduce significant performance degradation. 

\subsubsection{Retrieval Unification}

For a given question, there are multiple search vectors, based on the number of generated hypothetical tools, formatted according to a particular strategy (\texttt{TND}/\texttt{QTND}). Thus, the retriever returns multiple top-K tool lists, one for each vector. As not all tools in these top-K lists are useful, we need to filter out non-important ones and unify the useful ones into a single list for the LLM. One straightforward way is to select the top-1 tool from each retrieved list. However, as these lists are generated for the same question, they might have dependencies and overlaps that such simple methods do not take into consideration. Thus, to combine each top-K retrieved tool list into a single one, we propose using \textbf{Reciprocal Rank Fusion} (\textbf{RRF}) \cite{10.1145/1571941.1572114}. RRF is a widely adopted technique \cite{10.1145/3726302.3730157, kuo-etal-2025-mmlf, 10.1145/3704268.3742687} to unify multiple ranked lists. It works on the principle of creating a ranked list that takes into account the relative position of items across each list. In other words, if an item in each list is consistently ranked high, it will be reflected in the final list also. 

RRF is traditionally used to unify ranked lists from different retrievers for the same query. However, in our case, we use different search vectors with corresponding retrieved lists for unification. This is thus beyond the original scope of RRF. However, we view the different search vectors as being connected to the \textit{same input query}. As such, we loosely interpret RRF for our framework as simply an approach to unify multiple lists that are related to the original query. Moreover, using methods such as selecting the top-1 tool from each retrieved list would yield a total number of tools, the same as the number of generated HT's, which would be inconsistent with evaluation (\S \ref{sec:experiments}). This concludes the inference phase of \textit{ToolDreamer}. 

\section{Experiments}
\label{sec:experiments}

In this section, we conduct experiments to evaluate the effectiveness of the proposed framework. We aim to answer four research questions (RQ): (i) the benefit/performance gains obtained by using our framework over current approaches; (ii) the relative importance of each component of our framework; (iii) whether our method is flexible, i.e., independent of the choice of underlying models and (iv) additional considerations for usage in production.

\newcolumntype{Y}{>{\centering\arraybackslash}X} 

\begin{table*}[t]
\centering
\scriptsize
\setlength{\tabcolsep}{1.2pt} 
\begin{tabularx}{\textwidth}{c l *{16}{Y}}
\toprule
\multirow{3}{*}{\begin{sideways}Train?\end{sideways}} 
 & Split & \multicolumn{4}{c}{Web} & \multicolumn{4}{c}{Code} & \multicolumn{4}{c}{Customized} & \multicolumn{4}{c}{Avg.} \\
\cmidrule(lr){3-6}\cmidrule(lr){7-10}\cmidrule(lr){11-14}\cmidrule(lr){15-18}
 & Model & N@10 & P@10 & R@10 & MRR & N@10 & P@10 & R@10 & MRR & N@10 & P@10 & R@10 & MRR & N@10 & P@10 & R@10 & MRR \\
\midrule
              & BM25 (0s)   & 27.25 & 5.72 & 36.25 & 28.10 & 29.50 & 4.90 & 37.38 & 29.83 & 31.90 & 8.70 & 36.50 & 39.63 & 29.55 & 6.44 & 36.71 & 32.52 \\
\(\times\)    & BM25 (TND) (\textbf{K})  & 28.39 & 6.31 & 39.85 & 28.29 & 33.91 & 6.47 & 47.15 & 32.25 & 27.61 & 8.41 & 35.34 & 31.31 & 29.97 & 7.06 & 40.78 & 30.62 \\
              & BM25 (QTND) & 31.41 & 6.82 & 42.83 & 31.72 & 36.50 & 6.51 & 49.34 & 34.81 & 33.48 & 9.13 & 38.88 & 40.72 & \textbf{33.80}\impr{+14\%} & \textbf{7.49}\impr{+16\%} & \textbf{43.68}\impr{+19\%} & \textbf{35.75}\impr{+10\%} \\
\midrule
              & Qwen3 (0s) & 35.81 & 7.20 & 47.09 & 36.64 & 39.58 & 6.54 & 51.32 & 38.53 & 39.89 & 10.65 & 47.70 & 46.14 & 38.43 & 8.13 & 48.70 & 40.44 \\
\(\times\)    & Qwen3 (TND) (\textbf{K}) & 35.94 & 7.93 & 50.15 & 35.5 & 42.44 & 7.86 & 58.35 & 40.05 & 38.12 & 10.53 & 49.14 & 41.82 & 38.83 & \textbf{8.77}\impr{+8\%} & \textbf{52.55}\impr{+8\%} & 39.12 \\
              & Qwen3 (QTND) & 37 & 7.55 & 48.93 & 37.68 & 45.3 & 7.76 & 58.21 & 43.90 & 41.56 & 10.82 & 50.04 & 47.47 & \textbf{41.29}\impr{+7\%} & 8.71 & 52.39 & \textbf{43.02}\impr{+6\%} \\
\midrule
              & COLT (Phase-1) &   33.27 & 6.06 & 38.8 & 36.15 & 8.66 & 1.37 & 11.87 & 7.92 & 23.41 & 6.21 & 26.67 & 29.55 & 21.78 & 4.55 & 25.78 & 24.54 \\
              & COLT (Phase-2) &   0.2   & -    & 0.21 & -     & 0.17 & -    & 0.31  & -    & 0.42  & -    & 0.56  & -     & 0.26  & -    & 0.36 & -\\
              & Qwen3 (Q) (\textbf{TR})    & 38.15 & 7.80 & 49.91 & 39.07 & 39.29 & 6.60 & 50.84 & 38.28 & 42.17 & 11.30 & 49.80 & 48.86 & 39.87 & 8.57 & 50.18 & 42.07 \\
\ding{51}     & Qwen3 (TND)  & 37.27 & 8.2 & 51.23 & 36.98 & 42.39 & 8.03 & 58.27 & 39.65 & 39.71 & 11.25 & 50.87 & 43.49 & 39.79 & 9.16 & 53.46 & 40.04 \\            
              & Qwen3 (QTND) & 39.10 & 8.28 & 51.99 & 39.50 & 45.93 & 8.18 & 59.51 & 44.23 & 42.38 & 11.55 & 51.27 & 47.90 & \textbf{42.47}\impr{+7\%} & \textbf{9.34}\impr{+9\%} & \textbf{54.26}\impr{+8\%} & \textbf{43.88}\impr{+4\%} \\
\bottomrule
\end{tabularx}
\caption{Performance of BM25 and Qwen3 using \textit{ToolDreamer}. Best scores are in bold. Percentage improvements in teal w.r.t either zero-shot (using questions only) performance (testing only) or when training on question-tool mapping (Qwen3). Q = Trained on question-tool mapping (\texttt{ToolRet} (\textbf{TR}) baseline); TND = Thought + Name + Descriptions; QTND = Question + TND; 0s = zero-shot. The training script provided by \citet{10.1145/3627673.3679847} for phase-2 of COLT only reports R@10/N@10. For each split, we only use their associated tools when performing retrieval. Averages are computed using each split's performance.}
\label{tab:ToolDreamer}
\end{table*}

\subsection{Experiment Settings}

\noindent \textbf{Dataset.} We evaluate \textit{ToolDreamer} on \texttt{ToolRet} \cite{shi-etal-2025-retrieval}, a comprehensive benchmark that combines 26 (35 with subsets) existing tool-calling datasets such as UltraTool \cite{huang-etal-2024-planning-creation} and Gorilla \cite{patil2024gorilla}. \texttt{ToolRet} has three high-level subsets. (i) \textbf{Web}: APIs in OpenAI JSON format to perform various database or file operations, e.g., \texttt{check\_account\_functionality} (\textit{Verify the functionality of the account to confirm login status}); (ii) \textbf{Code}: Functions related to ML workflows, etc., e.g., \texttt{identify\_visible\_attributes} (\textit{Identify the visible attributes of an object in an image}); (iii) \textbf{Customized}: Tools to enable a broad range of daily tasks, e.g., \texttt{convertCurrency} (\textit{Convert currency values from one currency to another}). After filtration (App. \ref{sec:filtration}), web has \textasciitilde 37K tools/\textasciitilde 5K queries; code has \textasciitilde 4K tools/\textasciitilde 2K queries; customised has $>$ \textasciitilde 3K tools/\textasciitilde 1K queries. The training split of \texttt{ToolRet} consists of \textasciitilde 200K examples from which we sample 5K instances to optimise for API cost and data quality (App. \ref{sec:filtration}).

\noindent \textbf{Evaluation Metrics.} We evaluate retriever quality using 4 standard IR (Information Retrieval) metrics. These are (i) \textbf{NDCG@K}/\textbf{N@K} (Normalized Discounted Cumulative Gain): evaluates the \textit{quality} of a ranked list by rewarding relevant items being placed at the top of the list and penalizing them being placed towards the bottom; (ii) \textbf{P@K} (Precision): The fraction of retrieved items that are relevant to the search; (iii) \textbf{R@K} (Recall): The fraction of relevant items that were retrieved, and (iv) \textbf{MRR} (Mean Reciprocal Rank): Inverse rank of the first relevant item in the retrieved list. For each measure, \texttt{@K} means the number of items that are retrieved. Following \texttt{ToolRet}, we set K to 10.


\noindent \textbf{Baselines.} In \S \ref{sec:related}, we mention various related approaches for tool-retriever training. However, \citet{xu-etal-2024-enhancing-tool} does not have publicly available code. As such, we use (i) \citet{shi-etal-2025-retrieval} (\texttt{ToolRet} (\textbf{TR})), who train their models with both query-tool and (query + instruction) - tool similarity. However, we intentionally avoid incorporating the instructions to ensure a consistent evaluation with existing benchmarks which lack such annotations as our goal is to evaluate \textit{ToolDreamer} under the minimal assumptions that broadly apply across tool datasets; (ii) \texttt{COLT} \cite{10.1145/3627673.3679847}, a complex two-step framework that first tunes a model (\texttt{Contriever} \cite{izacard2022unsupervised}) using standard InfoNCE with query-tool similarity, followed by further optimisation with a graph neural network (GNN) using various decompositions of the training dataset; (iii) use HT directly for retrieval similar to \citet{kachuee-etal-2025-improving} (\textbf{K} in Table \ref{tab:ToolDreamer}).

\noindent \textbf{HT Generation Model.} We use GPT-4.1 \cite{openaiIntroducingGPT41} to generate hypothetical tools for both training and test data queries. Additionally, we show that \textit{ToolDreamer} can also use open-source models for HT generation (\S \ref{sec:Varying_HT_Generator}). Thus, it is not restricted by the choice of LLM used.

\subsection{\textit{ToolDreamer} Performance Analysis (RQ1)}
\label{sec:tooldreamer}

To evaluate the tool retrieval performance of \textit{ToolDreamer}, we choose a dense Qwen3-8B and a sparse BM25 model (App. \ref{sec:app_benchmarking}). Table \ref{tab:ToolDreamer} shows the results of \textit{ToolDreamer} against related baselines.

We evaluate our framework under training (using aligned tools) and no-training (zero-shot) settings. For the latter, it means directly feeding the search vectors (of the test set) to the models, using \texttt{TND}/\texttt{QTND} format, to gauge how well they work out-of-the-box with an optimised search component. As we can see from Table \ref{tab:ToolDreamer} (rows 1-6), \textit{ToolDreamer} offers strong improvements for both models, ranging between 8\% (Qwen3) and 19\% (BM25) over their zero-shot results. This highlights the effectiveness of our method, and particularly the quality of the generated hypothetical tools, which help the models acquire better cues on what tools to retrieve. We notice that while the plain \texttt{TND} setting works well, greater improvements are generally reported by including the question in the search vector. This reaffirms our hypothesis that hypothetical tools alone are insufficient for retrieval (Fig. \ref{fig:motivation}). While they provide an important signal, they benefit from coupling with the user query.


The last three rows of Table \ref{tab:ToolDreamer} show the impact of \textbf{training} Qwen with only questions (baseline from \citet{shi-etal-2025-retrieval}) v/s our aligned tools. Again, we see the benefit of our method in offering improvements. These results highlight two points: (i) using questions alone to train retrievers is sub-optimal, and (ii) when augmented with LLM reasoning chains, retrievers can be trained to better hone in on the required tools. It should be noted here that we use slightly fewer samples to achieve these scores as compared to training on questions alone. This is because, as mentioned in sec. \ref{sec:train_hyp_tool_gen}, we remove those questions from our sampled training dataset for which GPT is unable to generate the requested number of tools. This observation further showcases the quality of our generated tools and the sample efficiency of our framework. We provide an example of generated tools in App. \ref{sec:app_hyp_tool_quality}.

Next, we see how strong our framework is against \texttt{COLT}, another related baseline. Firstly, the results from phase-1 training (general query-tool alignment) are far superior to phase-2 (GNN-optimisation). This makes sense as they train their model using the standard query-tool InfoNCE objective, which works relatively well. However, the increased complexity of grouping tools, aligning them with queries, and using a GNN for further optimisation plummets performance. In this regard, \textit{ToolDreamer} is a much simpler and scalable approach capable of handling a variety of questions.

Finally, we would like to clarify that in this work, we intentionally exclude the instruction field of \texttt{ToolRet} dataset with the aim of developing/evaluating \textit{ToolDreamer} without assuming any a priori instruction directives.  Note that most existing tool-calling datasets lack instruction field annotations and annotating query instructions for new real-world use cases can be resource-intensive (e.g., requiring human experts to manually craft seed instructions as in \texttt{ToolRet}).  Without dependency on pre-exisitng instructions, \textit{ToolDreamer} can be readliy applied to other datasets as well as to new tool-calling use cases.



\subsection{Ablation Studies (RQ2)}

We conduct various ablation studies to investigate the importance of each step of \textit{ToolDreamer}: (i) the impact of HT quality (\S \ref{sec:ab_poor_quality_tools}) (ii) using a weak discriminator for tool alignment along with the alignment algorithm itself (\S \ref{sec:weak_tool_alignment}), and (iii) whether using an LLM for ranked list fusion offers additional benefits or not (\S \ref{sec:LLM-F}).

\subsubsection{Hypothetical Tool Quality}
\label{sec:ab_poor_quality_tools}

The foundation of \textit{ToolDreamer} starts with the reasoning ability of an LLM, and the subsequent quality of generated HT's. Thus, it is expected that \textit{higher quality HT's} equate to \textit{better downstream performance}. To investigate this, we tasked GPT-4.1 to generate \textit{inferior} quality tools by removing key requirements from the original prompt, using looser language and not providing any guiding examples (Fig. \ref{fig:ht_ablation_prompt}). By deliberately redacting guidance, we can \textbf{gauge the importance of writing well-crafted prompts} to generate higher quality HT's. Results from this test are shown in Table \ref{tab:poor_tools_ab}. As we can see, each model suffers from performance degradation when using inferior quality tools. Although a decrease of 2\% might not seem substantial, recent studies such as \citet{Yazan2024TheIO} indicate that even as little as two irrelevant results can have a catastrophic impact on the LLM's performance. As such, when using our framework, it is essential to generate quality HT's for augmenting the search vectors.

\begin{table}[t]
\small
\setlength{\tabcolsep}{5pt}
\centering
\begin{tabular}{@{}cccccc@{}}
\toprule
Model                  & Tools & NDCG@10 & P@10 & R@10  & MRR   \\ \midrule
\multirow{2}{*}{BM25}  & Original & \textbf{29.97}   & \textbf{7.06} & \textbf{40.78} & 30.62 \\
                       & Ablation & 29.55\decr{-1\%}   & 6.77\decr{-4\%} & 39.66\decr{-3\%} & \textbf{30.76} \\
\midrule
\multirow{2}{*}{Qwen3} & Original & \textbf{38.83}   & \textbf{8.77}  & \textbf{52.55} & 39.12  \\
                       & Ablation & 38.23\decr{-2\%}   & 8.33\decr{-5\%} & 50.58\decr{-4\%} & \textbf{39.21} \\ \bottomrule
\end{tabular}%
\caption{Impact of using poor-quality tools on performance. Best scores are in bold. Scores shown are averages from each split. Each setting uses the \texttt{TND} format to isolate the tool quality impact more effectively.} 
\label{tab:poor_tools_ab}
\end{table}

\subsubsection{Weak Tool Alignment}
\label{sec:weak_tool_alignment}

The second training step of our framework is tool alignment, where we create pairs of (hypothetical, gold) tools, based on their semantic similarity and assignment using a graph matching algorithm. This step is essential as poor HT-GT alignment can hamper the retriever's ability to learn meaningful relationships between the tools and the related question. As such, we examine two aspects: (i) quality of the embedding model used to create tool representations and (ii) graph matching algorithm, i.e., the method used to align the generated tool embeddings. These tests reveal the impact of using quality embeddings for alignment - highly similar representations for dissimilar tools (poor embedding model) can cause the alignment method to make incorrect assignments. Additionally, a weaker alignment technique can get stuck in a local optimum and ignore the global best assignment. To test each component, we (i) swap out Qwen3 embeddings for the far inferior DPR model \cite{karpukhin-etal-2020-dense} but use the Hungarian algorithm and (ii) use Qwen3 embeddings with a \textit{greedy} matching where at each step, it chooses the tool pair with the highest similarity. This method of selection is similar to greedy decoding in LLMs \cite{song-etal-2025-good}.

Results from this test are shown in Table \ref{tab:poor_alignment_ab}. Interestingly, while there is technically a decrease, it is not as strong as using poor quality tools (\S \ref{sec:ab_poor_quality_tools}). This indicates that while using a good embedding model/alignment method is necessary, even more so is having high-quality tools. These results suggest that even if there exists HT-GT misalignment, the very presence of the proposed external signal (HT) is enough to enhance retriever performance.

\begin{table}[t]
\centering
\small 
\setlength{\tabcolsep}{0pt} 
\begin{tabularx}{\columnwidth}{l *{4}{Y}} 
\toprule
Alignment Method & N@10        & P@10          & R@10           & MRR            \\ \midrule
Original         & \textbf{39.79} & \textbf{9.16} & \textbf{53.46} & \textbf{40.04} \\
DPR/Hungarian    & 39.48\decr{-1\%} & 9.10\decr{-.7\%} & 53.0\decr{-.9\%} & 39.81\decr{-.6\%} \\                
Qwen/Greedy      & 39.70\decr{-.2\%} & 9.15\decr{-.1\%} & 53.22\decr{-.4\%} & 40.0\decr{-.1\%} \\ \bottomrule

\end{tabularx}%
\caption{Impact of different alignment methods (embedding/matching algorithm) on training Qwen3. Base scores are provided for easier comparison. Each setting uses the \texttt{TND} format to isolate the tool-alignment quality impact more effectively.}
\label{tab:poor_alignment_ab}
\end{table}

\subsubsection{LLM List Fusion}
\label{sec:LLM-F}

Using LLMs to rank/rerank a top-K list of items \cite{10.1145/3696410.3714922, gangi-reddy-etal-2024-first, zhang-etal-2025-enhancing} has emerged as a novel solution to existing fusion mechanisms due to their superior reasoning capabilities. As such, we wanted to see how swapping out RRF for LLM-based fusion impacts overall retrieval performance. 

Here, we provide the set of retrieved tools for a question and ask (Fig. \ref{fig:LLM-F}) the LLM (GPT-4.1) to rank the top-10 in decreasing order of relevance with a score between 0 (least likely) and 1 (highly relevant). There are two cases to deal with: (i) when the number of retrieved tools = 10 (either using the question as the search vector or 1 HT) - the LLM is asked to rerank it based on query relevance, and (ii) number of tools > 10 (multiple HT's) - the LLM selects the top-10 among the entire set. 

Table \ref{tab:LLM-F} shows the results of this test. As seen, LLM reranking provides additional improvements over RRF. This makes sense as RRF is a simple statistical technique that simply considers the overall ranking of elements across lists, whereas an LLM does deeper reasoning to figure out which tools actually make sense for a query. However, these results come with caveats such as additional API costs, ignoring instructions, and hallucinations (spurious generations) \cite{10.1145/3703155}, which we elaborate in App. \ref{sec:app_LLM-F_issues}. Considering these issues, we prioritise RRF over LLM-ranking in \textit{ToolDreamer} as it offers reproducible and deterministic guarantees. That said, it is easy to equip \textit{ToolDreamer} with LLM-ranking, should cost/instruction-refusal not be a pressing concern.


\begin{table}[t]
\small
\begin{tabular}{@{}ccccc@{}}
\toprule
Approach   & NDCG@10        & P@10           & R@10           & MRR            \\ \midrule
TND/RRF    & 39.79          & 9.16           & 53.46          & 40.04          \\
QTND/RRF   & 42.47          & 9.34           & 54.26          & 43.88          \\ \midrule
TND/LLM-F  & 46.48          & \textbf{10.16}\impr{+9\%} & 57.64          & 46.83          \\
QTND/LLM-F & \textbf{46.53}\impr{+10\%} & 10.14          & \textbf{57.67}\impr{+6\%} & \textbf{46.89}\impr{+7\%} \\ \bottomrule
\end{tabular}
\caption{Impact of using different ranked-list fusion methods. LLM-F = LLM (GPT-4.1) Fusion. All scores are from the trained Qwen3 model using TND/QTND vectors. Improvements are shown against the best baseline (QTND).} 
\label{tab:LLM-F}
\end{table}

\subsection{Varying HT Generator (RQ3)}
\label{sec:Varying_HT_Generator}

In our experiments, we use GPT-4.1 to create our HT's. However, this might be a bottleneck in cases where API-calling might be an issue, either due to cost or privacy concerns \cite{yao2024survey}. Thus, we wanted to see how switching out GPT-4.1 for an open-source model impacted performance. To this end, we use Qwen3-32B (with the same prompt (Fig. \ref{fig:ht_test_prompt}) as GPT-4.1) as the HT generator and BM25 and NVIDIA \texttt{NV-Embed-2} \cite{lee2025nvembed} as the retrievers. Each model is evaluated with inference mode (no training) using both \texttt{TND/QTND} settings. A different dense retriever is used here to gauge the universality of our approach. We use the code split of \texttt{ToolRet} for these tests. The results are presented in Table \ref{tab:qwen-tools}. As we see, \textit{ToolDreamer} is flexible enough to work with different LLMs without much loss of performance. Thus, there is no necessity to stick to API models for HT generation, which shows the cost-benefit of using \textit{ToolDreamer}.

\begin{table}[t]
\centering
\footnotesize
\setlength{\tabcolsep}{4pt}
\begin{tabular}{llcccc}
\toprule
Model & Setting & N@10 & P@10 & R@10 & MRR \\ 
\midrule
\multirow{4}{*}{\hspace{5pt}\raisebox{-0.5\totalheight}{\rotatebox{90}{BM25}}}
 & TND / GPT & 33.91 & 6.47 & 47.15 & 32.25 \\
 & QTND / GPT & 36.50 & 6.51 & 49.34 & 34.81 \\
 & TND / Qwen & 31.43 & 5.52 & 42.68 & 30.19 \\
 & QTND / Qwen & 35.27 & 6.00 & 47.47 & 33.68 \\ 
\midrule
\multirow{4}{*}{\hspace{5pt}\raisebox{-0.5\totalheight}{\rotatebox{90}{NV-Emb}}} 
 & TND / GPT & 44.52 & 8.22 & 60.31 & 42.61 \\
 & QTND / GPT & 47.41 & 8.07 & 60.76 & 46.53 \\
 & TND / Qwen & 43.13 & 7.35 & 56.57 & 42.42 \\
 & QTND / Qwen & 46.11 & 7.56 & 59.16 & 45.31 \\
\bottomrule
\end{tabular}
\caption{Impact of using different Hypothetical Tool Generators. Setting = Search Vector / HT Generator.}
\label{tab:qwen-tools}
\end{table}

\subsection{Production Considerations (RQ4)}
\label{sec:production_considerations}

As we report our main scores using an API-based model, it is necessary to understand the associated costs, as higher monetary requirements can make the framework unsuitable for real-time production demands. Implementing \textit{ToolDreamer} cost us less than \$5 using OpenAI’s batch mode and the cheaper GPT-4.1. Thus, we can claim that our framework is highly cost-effective.

To gauge wall-time, we consider the query \textit{Can you tell me about the historical events of April 21st?} from the dataset. Basic retrieval (Qwen) takes \textasciitilde 0.04 seconds. With \textit{ToolDreamer}, it takes us \textasciitilde 8 seconds to generate the hypothetical tools using Qwen3-32B (on 1 A100 card) and \textasciitilde 2.5 seconds with GPT. Semantic search + RRF takes an additional 0.5 seconds, bringing the total to \textasciitilde 3 - 8.5 seconds. While this does introduce some latency, it is unlikely that in production, clients will constantly be asking questions that require tool usage, which helps justify the time constraints.

\section{Conclusion}

In this paper, we present \textit{ToolDreamer}, a flexible tool-retrieval framework that proposes a new lens for search, i.e., conditioning retrievers on \textit{hypothetical tools} potentially useful for a query, instead of the query itself, for a more natural search alignment. Through extensive experiments, we highlight the benefits of our method over models that rely on traditional query-tool similarity. Having a better retriever is closely linked with LLM tool-calling ability. \textit{ToolDreamer} is a step in this direction to provide LLMs with scoped tool sets connected to user queries. Future work will explore alternative loss objectives and tool alignment algorithms to further improve retriever efficiency.

\section*{Limitations}

We identify two limitations with our framework and offer potential workarounds. First, \textit{ToolDreamer} requires \textit{high-quality hypothetical tools} to work well (\S \ref{sec:ab_poor_quality_tools}). This requires a few rounds of testing to standardise the prompt, by experimenting with format, instructions, and examples. However, considering the importance of prompt quality \cite{DBLP:conf/acl/LongDNKCJK25} for most generative tasks, we find this requirement reasonable for \textit{ToolDreamer}. Second, compared to regular retrieval, RRF adds another layer of processing to increase the evaluation time complexity. However, from our evaluations (\S \ref{sec:production_considerations}), we did not find this to be a significant bottleneck considering the capability of modern hardware.


\bibliography{bibliography}

\appendix



\section{Benchmarking Current Retrievers}
\label{sec:app_benchmarking}

We benchmark several popular retriever models to establish a baseline and determine the best-performing ones to improve on. The selection criteria for the models were based on their performance on the BEIR \cite{thakur2021beir} and MTEB \cite{enevoldsen2025mmteb} leaderboards, which are considered the gold standard for ranking retriever models. We categorise them as, 

\begin{itemize}
    \item \textbf{Sparse}: Models that rely on surface-level token-matching for search. We use the popular BM25 \cite{bm25s} for this category
    
    \item \textbf{Semi-Sparse}: These are neural models that produce high-dimensional vectors (mostly 0's) but \textit{learn} the non-zero elements to capture better semantics while optimising for search efficiency. We select SPLADE \cite{lassance2024spladev3} and opensearch \cite{geng2024towards} for this category

    \item \textbf{Dense}: Models that generate low-dimensional embeddings aimed at capturing deeper semantics over simple token matching. These include DPR, BGE \cite{10.1145/3626772.3657878}, Jina \cite{gunther2023jina}, MiniLM \cite{wang-etal-2021-minilmv2}, NV-Embed, LGAI-Embedding \cite{choi2025lgaiembeddingpreviewtechnicalreport}, QZhou-Embedding \cite{yu2025qzhouembeddingtechnicalreport} and Qwen3 \cite{qwen3embedding}. We only use one proprietary model, i.e., OpenAI's \texttt{text-embedding-3-small} \cite{oaiembedding} to maintain cost.
\end{itemize}


The results from benchmarking are shown in Table \ref{tab:benchmarking}. Firstly, we notice the competitiveness of sparse retrievers for tool-retrieval. This aligns with observations by related studies \cite{shen-etal-2024-retrieval, zhuang-etal-2024-promptreps} which show how simple term-matching based models generalise to different tasks, better than dense models. In our tests, we find BM25 and opensearch matching the performance of MiniLM and even \texttt{text-embedding-3-small}, which not only highlights their tool-retrieval capabilities but also as an effective solution for settings where training dense models is infeasible.

However, when scaling up model size and moving beyond lexical features, we find that retrieval performance improves dramatically. This is evident from the gap between the best sparse (opensearch) and best dense (Qwen3) model, showing as much as 20\% increase (for MRR). Thus, it is clear that while sparse encoders are reasonably effective, to observe true tool-retrieval capabilities, dense models are fundamentally necessary.

For testing \textit{ToolDreamer}, we choose BM25 and Qwen3 as the best sparse/dense models, respectively. While SPLADE and opensearch are categorised as \textit{semi-sparse}, they still involve some form of contrastive learning to represent tokens. As such, we wanted to see how our method performs when using a \textit{pure} lexical model. Improvements for BM25 would additionally indicate the quality of the generated tools, as it means that the LLM effectively generates descriptions similar in phrasing to the expected tools. 

\begin{table*}
\centering
\scriptsize
\setlength{\tabcolsep}{2.8pt}
\begin{tabular}{l l *{16}{c}}
\toprule
 & Split 
 & \multicolumn{4}{c}{Web} 
 & \multicolumn{4}{c}{Code} 
 & \multicolumn{4}{c}{Customized} 
 & \multicolumn{4}{c}{Avg.} \\
\cmidrule(lr){3-6} \cmidrule(lr){7-10} \cmidrule(lr){11-14} \cmidrule(lr){15-18}
 & Model 
 & N@10 & P@10 & R@10 & MRR 
 & N@10 & P@10 & R@10 & MRR 
 & N@10 & P@10 & R@10 & MRR 
 & N@10 & P@10 & R@10 & MRR \\
\midrule
\multirow{3}{*}{\begin{sideways}Sparse\end{sideways}} 
 & BM25 & 27.25 & 5.72 & 36.25 & 28.10 & 29.50 & 4.90 & 37.38 & 29.83 & 31.90 & 8.70 & 36.50 & 39.63 & 29.55 & 6.44 & 36.71 & 32.52 \\
 & SPLADE-v3 & 29.22 & 6.08 & 39.52 & 29.56 & 26.45 & 4.96 & 37.24 & 25.52 & 36.89 & 10.12 & 46.24 & 41.16 & 30.85 & 7.05 & 41.00 & 32.08 \\
 & opensearch-embedding & 31.97 & 6.63 & 42.64 & 32.32 & 27.58 & 5.10 & 38.58 & 25.45 & 38.32 & 10.69 & 46.62 & 43.54 & 32.62 & 7.47 & 42.61 & 33.77 \\
\midrule
\multirow{9}{*}{\begin{sideways}Dense\end{sideways}} 
 & DPR & 7.41 & 1.59 & 11.01 & 7.35 & 9.76 & 1.95 & 13.35 & 9.16 & 11.93 & 2.75 & 15.73 & 13.87 & 9.70 & 2.10 & 13.36 & 10.13 \\
 & BGE-Large & 29.16 & 5.97 & 39.46 & 29.66 & 30.40 & 5.52 & 42.18 & 28.15 & 34.86 & 9.47 & 41.94 & 41.10 & 31.47 & 6.99 & 41.19 & 32.97 \\
 & jina-v2~ & 28.34 & 5.51 & 37.40 & 28.97 & 28.41 & 5.02 & 37.90 & 27.90 & 35.23 & 9.46 & 44.47 & 39.91 & 30.66 & 6.66 & 39.92 & 32.26 \\
 & all-MiniLM-L6-v2 & 29.53 & 5.77 & 39.05 & 30.38 & 27.23 & 4.92 & 37.11 & 26.85 & 33.15 & 8.83 & 40.81 & 38.76 & 29.97 & 6.51 & 38.99 & 32.00 \\
 & NV-Embed-v2 & 33.27 & 6.59 & 43.65 & 33.80 & 38.79 & 6.66 & 52.05 & 36.60 & 41.31 & 10.60 & 50.60 & 47.10 & 37.79 & 7.95 & \textbf{48.77} & 39.17 \\
 & \texttt{text-embedding-3-small} & 32.72 & 6.42 & 42.47 & 33.30 & 26.86 & 5.00 & 37.19 & 25.15 & 38.01 & 9.94 & 44.96 & 44.24 & 32.53 & 7.12 & 41.54 & 34.23 \\
 & LGAI-Embedding & 32.63 & 6.47 & 43.08 & 32.82 & 34.02 & 5.69 & 44.01 & 33.55 & 37.23 & 9.90 & 45.69 & 42.77 & 34.63 & 7.35 & 44.26 & 36.38 \\
& QZhou-Embedding & 23.67 & 4.61 & 31.76 & 23.8 & 21.95 & 4.19 & 30.02 & 21.95 & 30.14 & 7.84 & 37.07 & 35.11 & 25.25 & 5.55 & 32.95 & 26.95 \\
 & \textbf{Qwen3-8B} & 35.81 & 7.20 & 47.09 & 36.64 & 39.58 & 6.54 & 51.32 & 38.53 & 39.89 & 10.65 & 47.70 & 46.14 & \textbf{38.43} & \textbf{8.13} & 48.70 & \textbf{40.44} \\
\bottomrule
\end{tabular}
\caption{Zero-shot retriever benchmarking on each split of \texttt{ToolRet}. Best scores are in bold.} 
\label{tab:benchmarking}
\end{table*}

\section{LLM Performance v/s Tool Number}
\label{sec:app_OOM}

\begin{figure}
    \centering
    \includegraphics[scale=0.3]{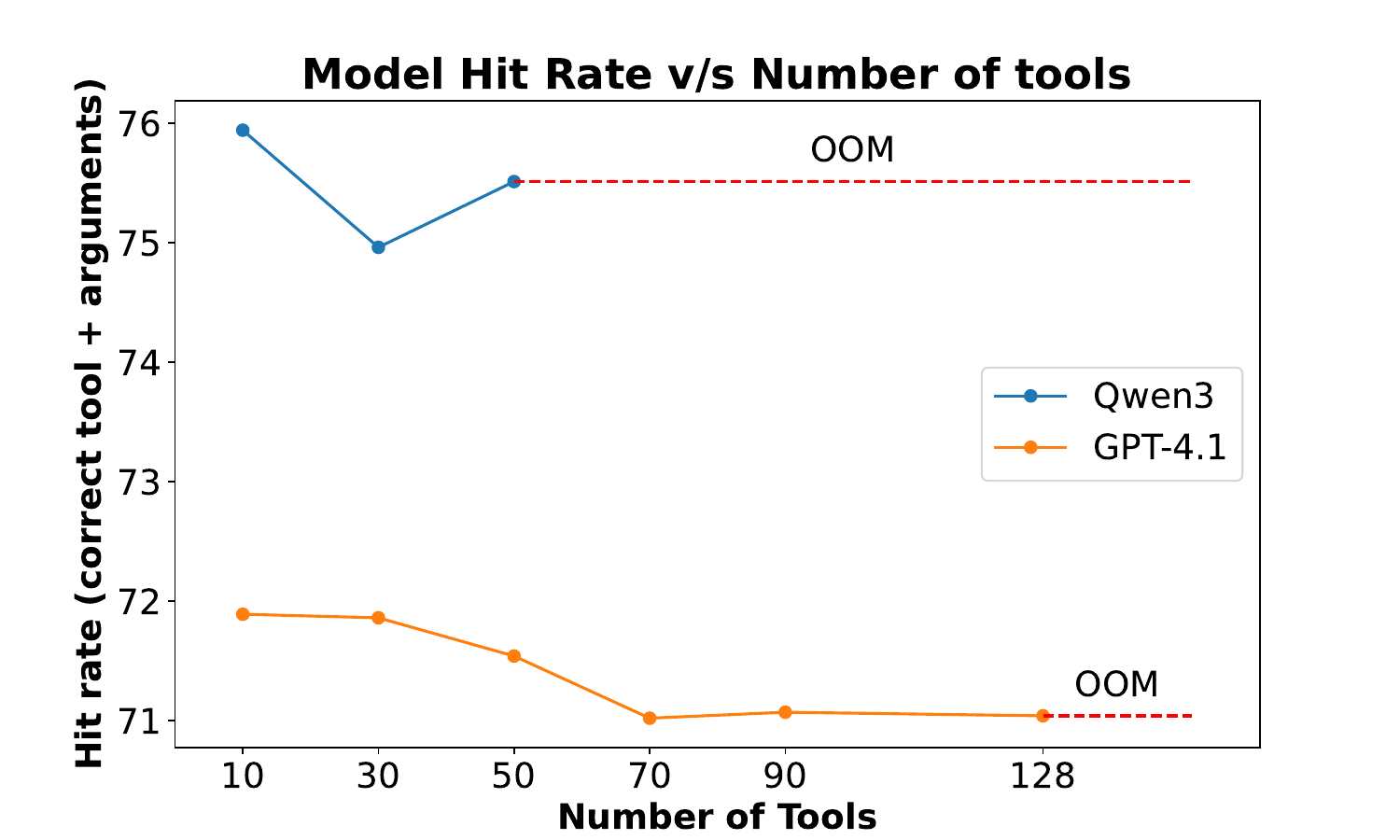}
    \caption{Impact of scaling the number of tools on LLM hit-rate (correct tool call with appropriate arguments).}
    \label{fig:qwen_vs_gpt_increasing_tools}
\end{figure}

To highlight the issue of tool limits, we examine how GPT-4.1 \cite{openaiIntroducingGPT41}, and Qwen3's \cite{yang2025qwen3technicalreport} performance changes as the number of tools increases (c.f. Fig. \ref{fig:qwen_vs_gpt_increasing_tools}). We use \texttt{SealTools} \cite{10.1007/978-981-97-9434-8_29}, a dataset for LLM-tool call training for illustration. For each query, we randomly sample \textit{negative} (irrelevant) tools to meet the tool counts (10-128). For example, if a query has 1 gold tool, we randomly sample 9 negative tools and shuffle their order to prevent location bias \cite{liu-etal-2024-lost}. As we can see, at relatively small scales (10-128 tools), each model handles tool calling relatively well, showing consistent performance across varying tool counts. However, beyond 50 (for Qwen3) and 128 (for GPT-4.1), each model hits their limit (Qwen: 32K context window; GPT: 1M), being unable to register more tools. GPT has another issue - since it is an API-based model, OpenAI/Azure restricts the maximum number of tools to 128 \cite{microsoftAzureOpenAI}. Thus, despite theoretically being able to support even more tools, given a 1M context window, GPT cannot be provided with more tools. These issues reinforce the necessity for tool retrieval.

\section{Training/Hardware Details}
\label{sec:training_details}

We train our models using LoRA \cite{hu2022lora} [rank=8, alpha=16, dropout=0.1] for 1 epoch, with a learning rate of $1e-5$ and batch size 1. These parameters were selected after brief experimentation with standard training values \cite{sbertTrainingOverview, huggingfaceLoRALowRank}, which showed to work the best. We run our training experiments on 4 NVIDIA A100 80 GB cards and inference on 1 of them.

\section{Note on LLM-Fusion (Re-Ranking)}
\label{sec:app_LLM-F_issues}

Here, we provide our justification for not using LLM-fusion in \textit{ToolDreamer}. First, using an LLM for ranking top-K lists requires additional API calls, which incur cost and can be prohibitive for a large number of queries. Second, LLMs are occasionally prone to missing instructions \cite{harada2025curse, DBLP:journals/corr/abs-2502-15851}, i.e., ignoring key requirements in the prompt. We notice a similar trend where for many samples, GPT under-reported (less than 10) or over-reported (more than 10) tools. For the former case, we use each generated tool, and for the latter, we retain the top 10. Although this setup works, it is slightly \textit{unfair} to compare it with RRF since it \textit{always} works with 10 tools impacting the metrics accordingly. Finally, although LLMs have become more reliable, the risk of \textit{hallucination}, i.e., spurious responses \cite{10.1145/3703155}, is not completely mitigated. As a result, LLM-based ranking \textit{may} introduce tools that are absent from the provided list, which is a major drawback. While hallucination concerns exist for hypothetical tool generation also, there we encourage open-ended creativity, which is detrimental for a systematic process such as list ranking.

\section{Dataset Filtration}
\label{sec:filtration}

To solve the retrieval problem, we need to ensure two things: (i) each tool must have a valid (non-empty) description to build the tool database index, and (ii) each positive/negative tool for a query must also have a valid description to perform retrieval. Accordingly, we filter the dataset to include tools and queries that satisfy these conditions. We found that the customised split had no noisy samples, and only the web and code split needed cleaning. This removed 264 tools/77 queries and 1 tool from each split, respectively, maintaining a count close to the original number of samples.

The provided training dataset has \textasciitilde 200K samples. However, many samples had profanities, repeated queries, negative tools appearing in the positive tools list, and questions with 0 negative tools (after removing the overlapping tools). To keep our data generation reasonable and devoid of such entries, we filter and sample 5K unique queries from the base set.

\section{Generated Hypothetical Tools}
\label{sec:app_hyp_tool_quality}

We provide an example of generated tools for a query from the Web split of ToolRet in Figure \ref{fig:generated_tool_ex}. As we can see, the query is quite dense and requires multiple tools to be solved. The associated gold tools take care of each aspect of the request. The generated hypothetical tools then align with each gold tool based on GPT's reasoning. These tools are not aligned in the figure. However, their language provides enough clues to the embedding model to draw precise similarity boundaries.

\begin{figure*}
    \centering
    \begin{tcolorbox}[colback=red!5!white,colframe=red!75!black,title=\textbf{Example of Generated Tool}]
    \texttt{Query: } \textit{I'm organizing a charity event and need some rewards for the attendees. Can you suggest some rewards available on the Rewards as a Service platform that I can offer as incentives? Also, fetch a random kitten image to include in the event promotion materials. Lastly, provide me with the order history for my account to see if I have any pending orders.}
    
    \hdashrule{\linewidth}{1pt}{2pt}   
    \texttt{Gold Tool 1:} \textcolor{magenta}{\textit{Fetches random kitten image!}}
    
    \texttt{Gold Tool 2:} \textcolor{magenta}{\textit{Get more information about all the orders placed on this customer and account}}
    
    \texttt{Gold Tool 3:} \textcolor{magenta}{\textit{Gets the list of rewards available for the platform}}
    
    \hdashrule{\linewidth}{1pt}{2pt} 
    \texttt{Hypothetical Tool 1:} \textcolor{olive}{\textit{Retrieves a catalog of rewards currently offered by the Rewards as a Service platform.}}
    
    \texttt{Hypothetical Tool 2:} \textcolor{olive}{\textit{Returns a random kitten image suitable for use in event promotion or other materials.}}
    
    \texttt{Hypothetical Tool 3:} \textcolor{olive}{\textit{Retrieves the order history, including pending and completed orders, for a specified account on the Rewards as a Service platform.}}
    \end{tcolorbox}
    \caption{An example of hypothetical tools for a given query.}
    \label{fig:generated_tool_ex}
\end{figure*}

\section{Prompts}
\label{sec:app_prompts}

Below, we provide the prompts used to generate the hypothetical tools for \textit{ToolDreamer}.

\begin{figure*}
    \centering
    \begin{tcolorbox}[promptbox={Prompt For Inferior Quality Tools}]
    You are an assistant that comes up with tools. The tools should be related in some way to solving a query. Just give a short explanation, a name, and a description for each tool. \\

    Output Format

    Thought: [say something about the tool] \\
    Tool Name: [give it a name] \\
    Tool Description: [say what it does]
    \end{tcolorbox}
    \caption{Prompt for generating inferior quality hypothetical tools for test data (ablation study).}
    \label{fig:ht_ablation_prompt}
\end{figure*}

\begin{figure*}
    \centering
    \begin{tcolorbox}[promptbox={Prompt For LLM Reranking}]
    You are given a \textbf{question} and a \textbf{set of tools}, each with a description. Your task is to evaluate the tools and rank them by how useful they are for answering the question.

    \textbf{Instructions}:

    \begin{enumerate}
        \item If there are \textbf{10 tools or fewer}, rank and score \textbf{all of them}.
        \item If there are \textbf{more than 10 tools}, consider the full set and return \textbf{only the top 10} ranked by usefulness.
        \item Use a \textbf{score between 0.00 and 1.00}, where:
            \begin{itemize}[label=\ding{72}]
                \item 1.00 = extremely useful / directly relevant
                \item 0.00 = not useful at all
            \end{itemize}
        \item Start with a \textbf{concise reasoning summary} (2–4 sentences maximum).
        \item Then, present the ranked list in the following format, sorted from most to least useful:
        \begin{verbatim}
```
Tool: [tool name] 
Score: [0.00–1.00]
```
        \end{verbatim}
    \end{enumerate}
    \end{tcolorbox}
    \caption{Prompt for LLM-reranking (ablation study).}
    \label{fig:LLM-F}
\end{figure*}

\begin{figure*}[htbp]
    \centering
    \begin{tcolorbox}[promptbox={Prompt For Training Data}, width=\textwidth]
        You are a \textbf{tool-crafting assistant}. Your job is to design a precise set of \textbf{hypothetical tools} that collectively solve a given user query. Each tool must behave as an \textbf{independent, stateless function} with full access to all structured data relevant to its subtask. Assume \textbf{no external internet access}. 
        
        \#\#\# TASK INSTRUCTIONS 
        \begin{enumerate} 
        
            \item \textbf{Query Breakdown} \\ 
            Analyze the user query step by step. Identify all explicit and implicit \textbf{subtasks} required to generate a complete answer. 
            \begin{itemize}[label=\ding{72}] 
                \item If multiple subtasks are tightly related, keep them separate unless they are inseparable. 
                \item If there appear to be fewer subtasks than the number of tools requested, dig deeper to uncover \textbf{hidden or supporting subtasks}. 
            \end{itemize} 
            
            \item \textbf{Tool Design} \\ 
            For \textbf{each subtask}, propose \textbf{exactly one tool} to handle that specific function. 
            
            \item \textbf{Design Principles} 
                \begin{itemize}[label=\ding{72}] 
                    \item Tools must be: \textbf{modular}, \textbf{reusable}, and \textbf{implementation-agnostic}. 
                    \item Use \textbf{general tool names} unless a domain-specific name is clearly justified (e.g., `weatherAPI\_California'). If specificity is used, explain why. 
                    \item Tools should focus on a \textbf{single responsibility} each. 
                \end{itemize} 
                
            \item \textbf{Naming Rules} 
                \begin{itemize}[label=\ding{72}] 
                    \item Choose either `camelCase' or `snake\_case' and stick with it throughout. 
                    \item Tool names must be \textbf{concise, descriptive, and aligned directly to the subtask}. 
                    \item Avoid overly broad names (e.g., `processData') or overly detailed names (e.g., `getSanFranciscoHourlyWeatherFromNOAA'). 
                \end{itemize} 
                
            \item \textbf{Tool Count Compliance} 
                \begin{itemize}[label=\ding{72}] 
                    \item Always produce \textbf{exactly the number of tools specified} in the query. 
                    \item If subtasks seem fewer than required, introduce additional implicit tools (e.g., parsing input, validating scope, formatting output). 
                \end{itemize} 
            \end{enumerate} 
        
        \#\#\# OUTPUT FORMAT \\
        For each tool, provide the following fields \textbf{in order}: 

\begin{verbatim}
```
Thought: <Explain what this subtask is and why this tool is needed>
Tool Name: <Concise, consistent name>
Tool Description: <Briefly describe what the tool does>  
```
\end{verbatim}
    
\#\#\# EXAMPLE 

QUERY: What is the average temperature in San Francisco? | TOOLS NEEDED: 1

\begin{verbatim}
```
Thought: The query requires access to temperature data for a specific location. A tool is needed to 
fetch weather data for San Francisco.
Tool Name: weatherAPI\_California
Tool Description: Retrieves current and historical temperature data for locations in California.
```
\end{verbatim}
    \end{tcolorbox}

    \caption{Prompt for generating hypothetical tools for training data.}
    \label{fig:ht_train_prompt}
\end{figure*}

\begin{figure*}[htbp]
    \centering
    \begin{tcolorbox}[promptbox={Prompt For Test Data}, width=\textwidth]
    You are a tool-crafting assistant. Your goal is to design a set of \textbf{hypothetical tools} that can collectively solve a given user query. These tools behave like \textbf{independent, stateless functions} and have access to \textbf{all relevant structured data} needed to complete their task. No external internet access is assumed.

    \#\#\# High-Level Strategy

    \begin{enumerate}
        \item \textbf{Break down} the query into distinct, logically separable \textbf{subtasks}.
        \item For \textbf{each subtask}, propose \textbf{exactly one tool} that performs \textbf{only that task}.
        \item Tools must be \textbf{modular, reusable, and generic} where possible (e.g., `getPopulationData'). If a subtask clearly requires \textbf{specialized or scoped} logic (e.g., region-specific), it's acceptable to use a \textbf{specialized tool}, but its name must reflect its limited scope (e.g., `weatherAPI\_California', `fetchStockData\_TechSector').
        \item All tools must be:
            \begin{itemize}[label=\ding{72}]
                \item \textbf{Implementation-agnostic}
                \item \textbf{Independent}
                \item \textbf{Named consistently} using `camelCase' or `snake\_case'
            \end{itemize}
        \item \textbf{Every query must result in at least one tool} being proposed.
    \end{enumerate}

\#\#\# Output Format

For each tool, respond using the following format:
\begin{verbatim}
```
Thought: [Explain why this tool is necessary and what subtask it addresses]
Tool Name: [Concise and consistent tool name]
Tool Description: [Describe what this tool does in one or two sentences, avoiding implementation 
details]
```    
\end{verbatim}

\#\#\# Example

\#\#\#\# User Query:

\begin{verbatim}
```
What is the average temperature in San Francisco?
```    
\end{verbatim}

\#\#\#\# Assistant Response:
\begin{verbatim}
```
Thought: To answer this query, we need to retrieve historical and/or current temperature data for a 
specific location in California.
Tool Name: weatherAPI\_California
Tool Description: Retrieves current and historical temperature data for locations in California.
```    
\end{verbatim}

\#\# Additional Tips
\begin{itemize}[label=\ding{72}]
    \item Do \textbf{not} suggest tool pipelines or combine subtasks into a single tool.
    \item If the query involves \textbf{data transformation} (e.g., averaging, filtering, ranking), propose a separate tool for that.
    \item Treat each tool as \textbf{stateless and composable}.
\end{itemize}
    \end{tcolorbox}

    \caption{Prompt for generating hypothetical tools for test data.}
    \label{fig:ht_test_prompt}
\end{figure*}

\section{Paper Code}

The code for this paper is publicly available at \url{https://github.com/boschresearch/ToolDreamer}.

\end{document}